\documentclass[10pt, a4paper]{article}
\usepackage{lrec-coling2024} 
\usepackage{natbib}
\usepackage{multibib}
\makeatletter
\def\@mb@citenamelist{cite,citep,citet,citealp,citealt,citepalias,citetalias}
\makeatother
\newcites{languageresource}{~}

\usepackage{graphicx}
\usepackage{tabularx}
\usepackage{soul}

\usepackage{titlesec}
\titleformat{\section}{\normalfont\large\bfseries\centering}{\thesection.}{1em}{}
\titleformat{\subsection}{\normalfont\small\bfseries\raggedright}{\thesubsection.}{1em}{}
\titleformat{\subsubsection}{\normalfont\normalsize\bfseries\raggedright}{\thesubsubsection.}{1em}{}
\renewcommand\thesection{\arabic{section}}
\renewcommand\thesubsection{\thesection.\arabic{subsection}}
\renewcommand\thesubsubsection{\thesubsection.\arabic{subsubsection}}

\usepackage{xcolor}
\usepackage{url}
\usepackage{hyperref}
 \definecolor{darkblue}{rgb}{0, 0, 0.5}
  \hypersetup{colorlinks=true, citecolor=darkblue, linkcolor=darkblue, urlcolor=darkblue}

\usepackage{xstring}

\usepackage{color}

\graphicspath{ {./images/} }
\usepackage{listings}
\usepackage{nameref}
\usepackage[normalem]{ulem} 

\title{Code Book for the Annotation of Diverse Cross-Document Coreference of Entities in News Articles}

\name{Jakob Vogel} 

\address{M.A. Digital Humanities \\
         Institute for Digital Humanities, Faculty of Philosophy, Georg August University of Göttingen \\
         jakob.vogel@stud.uni-goettingen.de \\
         }

\abstract{
This paper presents a scheme for annotating coreference across news articles, extending beyond traditional identity relations by also considering near-identity and bridging relations. It includes a precise description of how to set up Inception, a respective annotation tool, how to annotate entities in news articles, connect them with diverse coreferential relations, and link them across documents to Wikidata's global knowledge graph. This multi-layered annotation approach is discussed in the context of the problem of media bias. Our main contribution lies in providing a methodology for creating a diverse cross-document coreference corpus which can be applied to the analysis of media bias by word-choice and labelling.
\\ \newline }

\begin{document}

\maketitleabstract

\section{Introduction}
\label{intro}

\newcounter{ex}
\setcounter{ex}{0}
\newcommand{\ex}[1]{\refstepcounter{ex}\label{#1}}

\begin{figure}[!ht]
    \begin{center}
    \includegraphics[width=8cm]{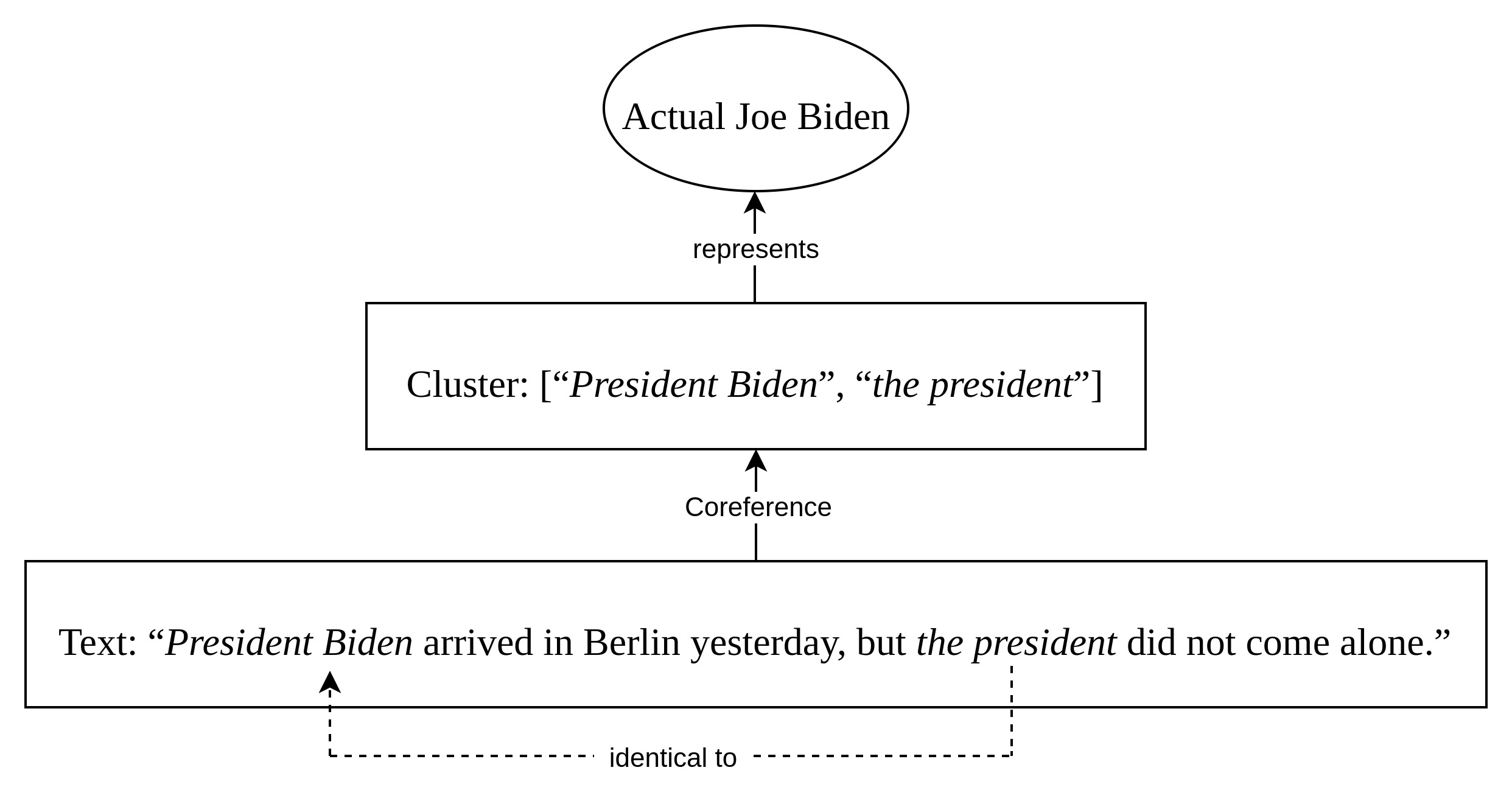}
    \caption{Illustration of how a cluster can be formed from an antecedent and its anaphor(s). The cluster represents its referent, in this case Joe Biden, in a text.}
    \label{fig:coref_basic}
    \end{center}
\end{figure}

Coreference is the phenomenon of several expressions in a text all referring to the same person, object, or other entity or event as their referent. Thus, in a narrow sense, analyzing a document with regards to coreference means detecting relations of identity between phrases. The following example (\ref{ID-1}) illustrates such an identity relation, where coreferential expressions are printed in italics:
\begin{quote}
\ex{ID-1}(\ref*{ID-1}) “\textit{Joe Biden} arrived in Berlin yesterday, but \textit{the president} did not come alone.”
\end{quote}
In (\ref{ID-1}), the noun phrase \textit{“Joe Biden”} introduces a new entity while \textit{“the president”} relates back to that introducing phrase. Within this relation, the introducing phrase \textit{“Joe Biden”} is called the \textbf{antecedent} while the back-relating phrase \textit{“the president”} is called an \textbf{anaphor}. Both expressions are coreferential in the way that they refer to the same non-textual entity, namely to the actual ‘real-world’ Joe Biden or at least to a corresponding mental concept. We can think of an antecedent and its anaphora as forming a \textbf{cluster} of \textbf{mentions} that as a whole represents its extra-textual referent within a textual document, as shown in Figure \ref{fig:coref_basic}.

As a task of natural language processing (NLP), coreference resolution has become quite efficient in detecting identity relations between phrases. However, reflecting on how we use language to refer to something, we are forced to realize that coreference in a broader sense is actually far more complex. We can address an entity or event by using a variety of expressions that are in fact not strictly identical to each other. Consider the following examples:
\begin{quote}
\ex{near-ID-1}(\ref*{near-ID-1}) “\textit{President Biden} was clearly not satisfied with today’s outcome. As \textit{the White House} stated this afternoon, efforts will be made to …”
\end{quote}
\begin{quote}
\ex{near-ID-2}(\ref*{near-ID-2}) “Even if \textit{the young Erdogan} used to be pro-Western, \textit{Turkey's president} nowadays often acts against Western interests.”
\end{quote}
\begin{quote}
\ex{near-ID-3}(\ref*{near-ID-3}) “The AfD is circulating \textit{a photo of Angela Merkel with a Hijab}, although \textit{Merkel} never wore Muslim clothes.”
\end{quote}
In these given examples, the highlighted mentions mean ‘almost’ the same, but not completely. In (\ref{near-ID-1}), we are aware by world-knowledge that "\textit{the White House}" is often used as a substitute expression for the current US president, although the former is a place which in strict terms cannot be identical to the president, who is a person. In (\ref{near-ID-2}), on the other hand, both mentions refer to the 'real-world' person Erdogan, but at different time steps. Finally, in (\ref{near-ID-3}), a mention representing the person Merkel is juxtaposed with a mention representing a picture of Merkel. While these two mentions could refer to separate entities, the juxtaposition indicates a connection between both where the attributes of the first mention do influence the perception of the second mention. Hence, we would miss essential semantic connections if we chose not to mark them as coreferential. Having said that, the simple classification of two mentions into coreferential (identical) or non-coreferential (non-identical) does not seem to suffice the complexity of common text data. Instead, we need to allow for \textbf{diverse coreference} clusters that include finer-grained relations lying between identity and non-identity. We need to allow for \textbf{near-identity} relations to mark two mentions that are partially, but not totally, identical \citep{recasens2010atypology}.

In news coverage, identity and near-identity references are extensively used to report on persons, organizations, and other entities of public interest. It is our goal to build up a corpus that contains annotated examples of such diverse forms of coreference. While diverse coreference occurs in all sorts of news media, we focus on digital print media, only. Furthermore, although in practice both entities and events can act as referent, we ignore references to events for now, as their annotation would go beyond the limits of our present scheme.\footnote{Though at a later point, this scheme could be extended to also include the annotation of events \citep{linguisticdataconsortium2005ace, o2016richer}.}

\begin{figure}[!ht]
    \begin{center}
    \includegraphics[width=8cm]{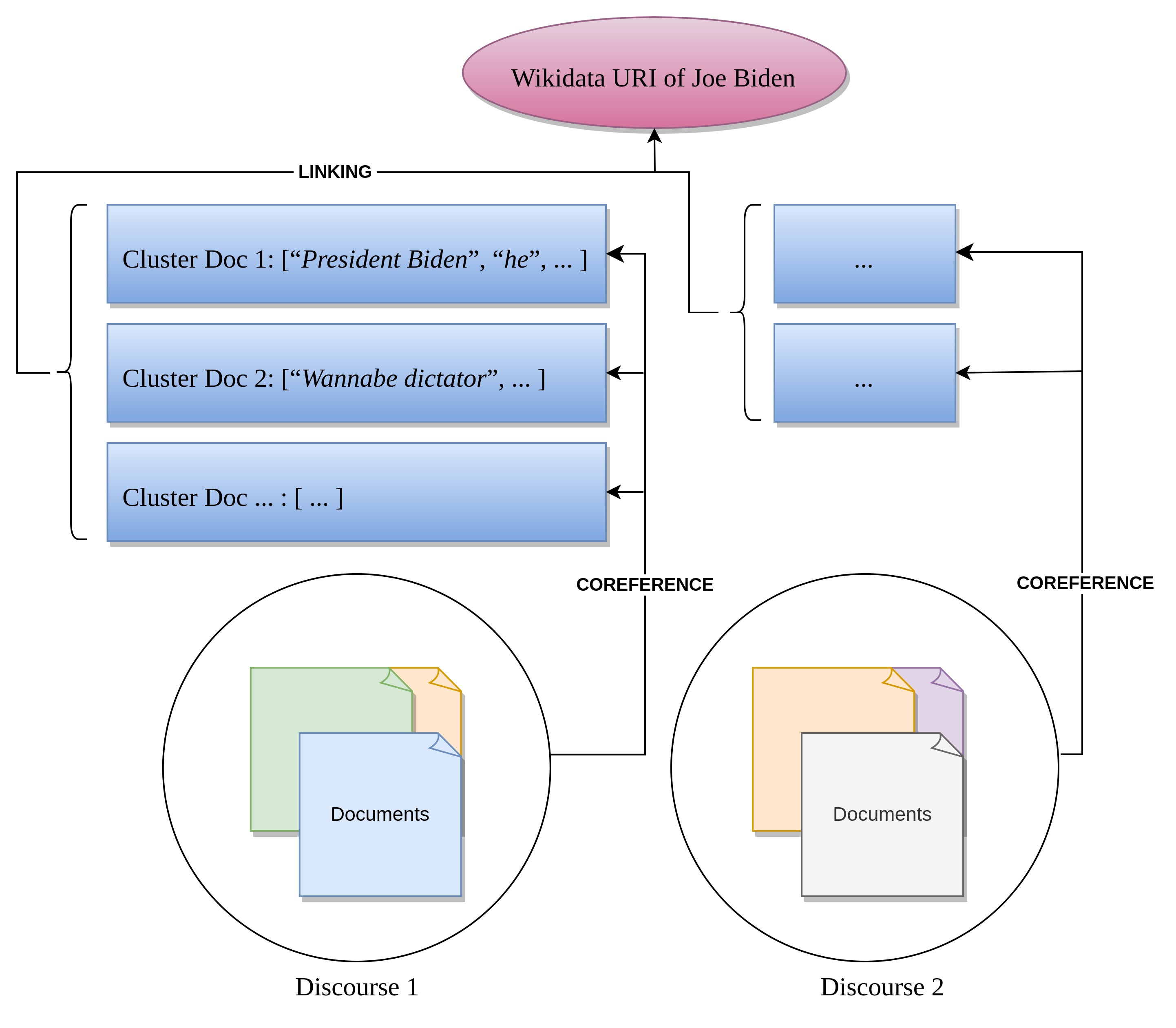}
    \caption{Illustration of our multi-layered annotation: within several discourses which all consist of multiple news articles reporting on the same happening, document entity clusters are extracted for each document. Those clusters are assigned a Wikidata URI. This ensures an unambiguous identification of each cluster, but it also links each cluster to all other clusters with the same referent within one discourse as well as across discourses. Finally, the linking also adds world knowledge to the annotated data.}
    \label{fig:multi-layer}
    \end{center}
\end{figure}

The ordinary business of journalism is to write about current political affairs and other happenings of public interest. These happenings are normally reported by several newspapers at the same time. All of these news articles are considered documents that contain references to the same entities and together form a discourse about them. To include the whole picture of such intradiscursive references, we want our corpus to link document-level clusters with corresponding clusters of other documents of the same discourse. Hence, our corpus is to depict \textbf{cross-document coreference} data. On a discourse level, corresponding clusters form discourse entities that themselves can be linked to their non-textual referents by some knowledge graph identifier. For this project, we use Wikidata's Uniform Resource Identifiers (URIs) for entity linking. By doing so, world knowledge is included into the data. This allows for drawing connections even between different discourse entities that refer to a common referent, yet at a different time step or rather in the context of a different happening. Figure \ref{fig:multi-layer} illustrates the multiple layers of this annotation model.

In building a corpus for diverse cross-document coreference in news articles, we hope to provide a valuable resource for the evaluation of automated coreference resolution tasks. The contribution of this paper mainly lies in providing an answer to the question of how to create such a corpus. How can diverse coreference relations be annotated in a cross-document setup? We believe our scheme as we present it here tackles this problem efficiently, extensively, and unambiguously.

Additionally, we would like to use the data resulting from our own annotations for further research in the area of media bias. Even if it plays no direct part in the outlined scheme, a lot of our choices how to annotate references were made because of this requirement to make the data usable for later media bias analysis. Eventually, we hope to contribute to the wider research question of how to identify media bias by word choice and labelling based on the usage of diverse coreference relations in news articles. 

The following section \ref{sec:media_bias} will further elaborate on this connection between diverse coreference and the problem of media bias analysis. Despite its only subtle impact on our practical annotation instructions, that section means to highlight the theoretical background and motivation behind our project. The sections thereafter will then deal with the actual annotation process. Section \ref{sec:tool} will guide coders through the setup and controls of Inception, our selected annotation software. Finally, section \ref{sec:annotation} will define annotation instructions in three passes while also outlining our typology of diverse coreference.

The data we use for our own annotations consists of the text bodies of articles that report on the same happenings. All articles are in English and were published by one of the following US-American newspapers: HuffPost (categorized as "Left" by AllSides \citeyearpar{allsides2023allsides} or "Skews Left" by Ad Fontes Media \citeyearpar{adfontesmedia2023interactive}, abbreviated in our data as "LL"), The New York Times (categorized as "Lean Left" by AllSides or "Skews Left" by Ad Fontes Media, abbreviated as "L"), USA Today (categorized as "Lean Left" or "Middle or Balanced Bias", abbreviated as "M"), Fox News (categorized as "Right" or "Skews Right", abbreviated as "R"), Breitbart News Network (categoized as "Right" or "Strong Right", abbreviated as "RR").\footnote{Looking at the political orientation of these newspapers, the data is unbalanced with an underrepresentation of politically centered media. However, at the current state of this project, the imbalance is unlikely to influence our analysis which does not yet target political orientation or media bias itself. Therefore, we will ignore this issue for now.}

\section{Diverse cross-document coreference and media bias analysis}
\label{sec:media_bias}

Media bias is a multifaceted phenomenon of news coverage that is one-sided, politically shaded, or in some other way non-neutral. It can occur in all sorts of news media, though we focus on digital print media, only. One specific type of media bias is \textbf{bias by word-choice and labeling} \citep{hamborg2019automated}. Word choice describes the selection from a variety of possible expressions to refer to an entity. For example, in order to refer to the USA’s current head of state, journalists could use one of the relatively neutral alternatives “Joe Biden”, “Biden”, or “the US president”, or in theory, choose a clearly biased expression like "the dictator" \citep{theguardian2023foxnews}.

Labeling, on the other hand, describes the assignment of attributes to an expression, inter alia by adding adjectives. Examples for bias by labeling include “an anxious and uncertain president” or “crooked Joe Biden” \citep{mediaite2023trump}.
Together, word-choice and labelling form a so-called \textbf{frame} \citep{hamborg2019automated}. 
In news articles, frames are used in a variety of ways, either for the sake of linguistic diversity or to make certain, potentially biased statements about an entity. To test an article for such statements, all of an entity's frames need to be extracted and evaluated together. Hence, before an article can be properly analyzed with regards to if and how it uses biased frames of (certain) entities, we are first faced with the task of identifying such frames. The identification of all expressions that refer to the same entity is a matter of coreference resolution. To conclude, successful coreference resolution is a prerequisite to any further inquiry of media bias by word-choice and labelling.

As already indicated above, automatic coreference resolution does show good results in extracting identity clusters from a document \citep{liu2023brief}. However, we have seen that there exist near-identity relations between expressions, potentially even across documents, that would be mostly overseen by standard coreference resolution approaches \citep{zhukova2022xcoref}. Hence, they would also be overseen by any media bias analysis that depends on coreference resolution. We hope that our building of a corpus for diverse cross-document coreference will contribute to the analysis of media bias by providing data that contains the full variety of frames used in news articles. Eventually, we would like to test how we can measure media bias by focusing on diverse coreference in news articles. To answer this last question, though, an additional layer of media bias annotation would have to be put upon our coreference data \citep{spinde2021you, spinde2021towards}. 

\section{Annotation tool}
\label{sec:tool}
The software we will use for annotation is called \textbf{Inception} \citep{klie2018theinception}. Inception is an open source annotation tool which can be freely downloaded from the authors' \href{https://inception-project.github.io/}{GitHub repository}. Although for this project, every annotator will be provided with a ready-to-code version of the program with all necessary annotation layers and settings already implemented and some sample annotations included. This instance of Inception can be requested from the project administrator \href{mailto:jakob.vogel@stud.uni-goettingen.de}{Jakob Vogel}.

\subsection{Setup}
To set up Inception on your local computer, make sure you already received your personal instance of the software. If not, please contact the project administrator.

Inception comes as a jar-file. In order to run it, you need to have the Java Runtime Environment (JRE) installed. Furthermore, make sure the file is set as executable. Then open the directory "Inception" in your command prompt and run: 
\begin{lstlisting}[language=bash, frame=single]
 java -jar inception.jar
\end{lstlisting}
To access Inception's graphical user interface (GUI), go to a web browser and open:\newline \url{http://localhost:8080/} \newline
On your first time running Inception, you will need to import the project and set up your personal user account:
\begin{itemize}
    \item First, log in as admin (User ID: \textit{admin} ; Password: \textit{admin}).
    \item Click on "Import project" and select the file "proj-div-CDCR.zip" from the "Inception" directory. Make sure to check the boxes "Import permissions" (already checked by default) and "Create missing users" (unchecked by default). Then click "Import".
    \item Click on "Administration" in the GUI's right top corner. Then click on "Users".
    \item Select your personal user or create a new one here. Assign a password to your user. Additionally, assign the role "ROLE\_USER" to your user (already assigned by default). Finally, check the box "Account enabled" and click "Save".
    \item Log out of the current Inception session.   
\end{itemize}
From now on, to log into Inception, use your personal user account details instead of the admin account.
\begin{figure}[!ht]
\begin{center}
\includegraphics[width=8cm]{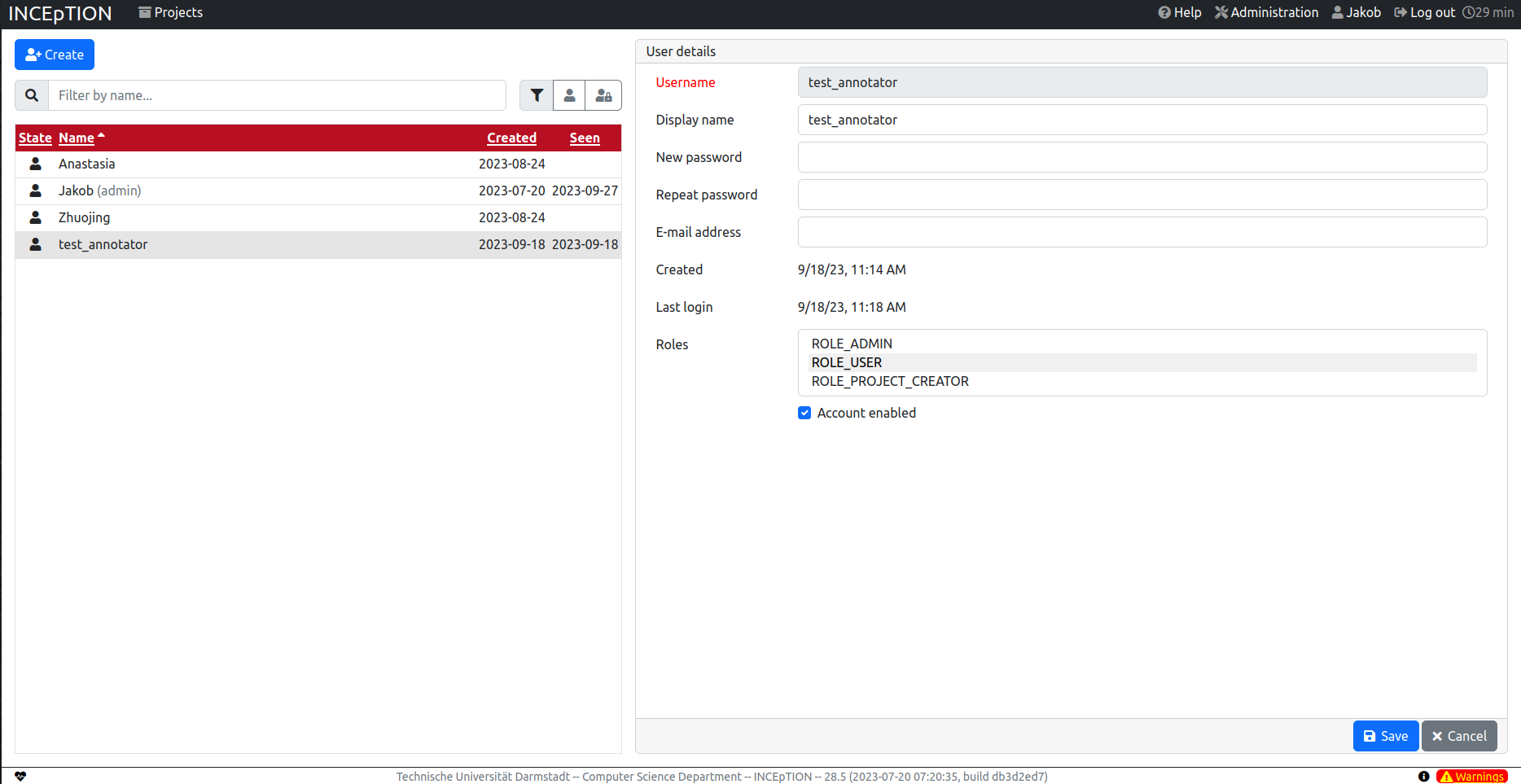}
\caption{Screenshot of Inception window showing the user management settings. Make sure to create or activate your own user account here at first login.}
\label{fig:inc10_users}
\end{center}
\end{figure}

To get to the annotation GUI, log in with your personal account now and click on the highlighted project name "Diverse cross-document coreference". Then, in the left taskbar, click on "Annotation". A window opens that shows a list of all documents to be annotated. The first digit in every title is a discourse identifier that sorts all documents according to their topic, followed by an underscore and a newspaper abbreviation (see \nameref{intro}). You can annotate documents in chronological order or randomly, whichever you prefer. Click on one of the documents to start your annotation.
\begin{figure}[!ht]
\begin{center}
\includegraphics[width=8cm]{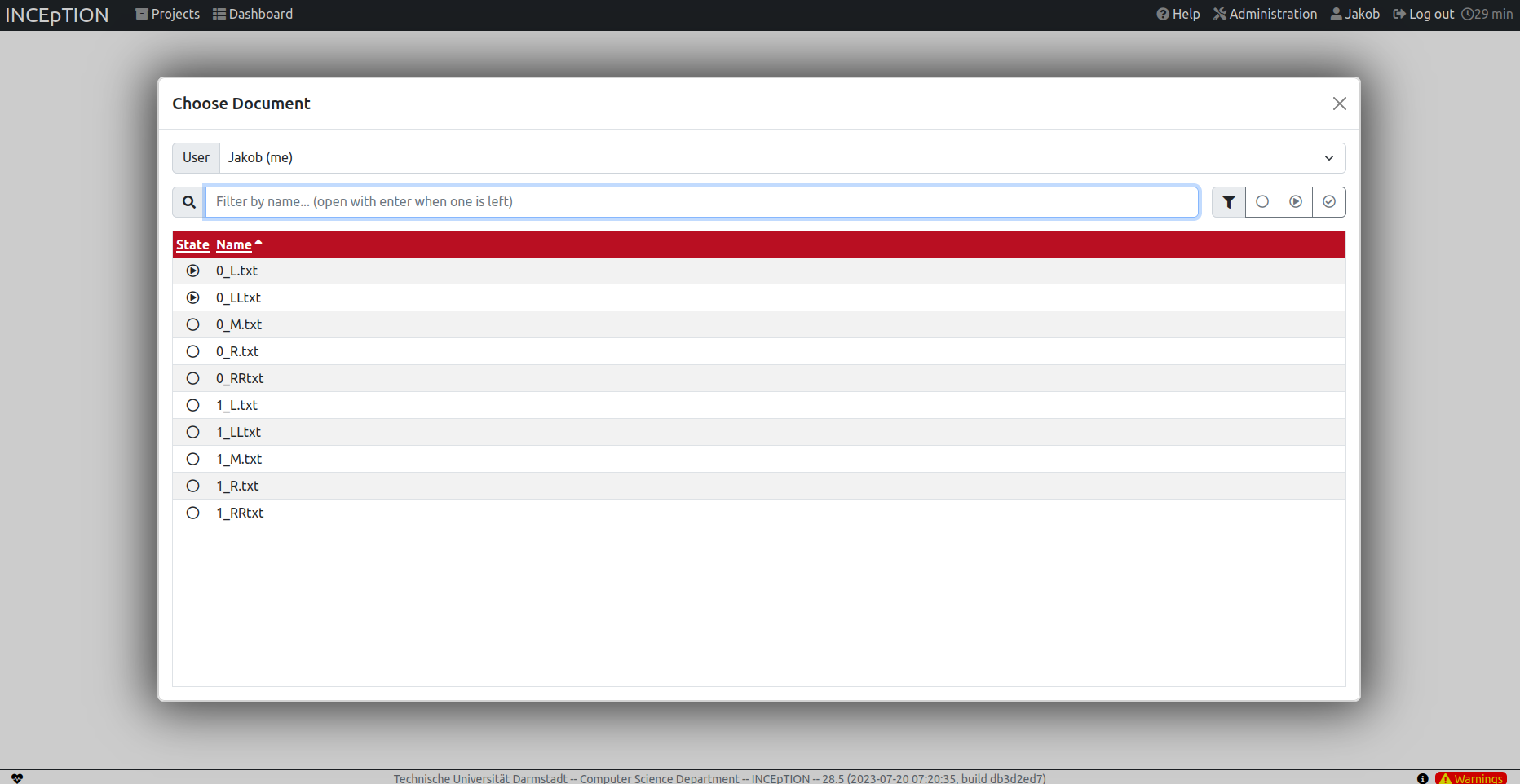}
\caption{Screenshot of Inception window showing a list of all documents to be annotated.}
\label{fig:inc4_documents}
\end{center}
\end{figure}

\begin{figure}[!ht]
\begin{center}
\includegraphics[width=\linewidth]{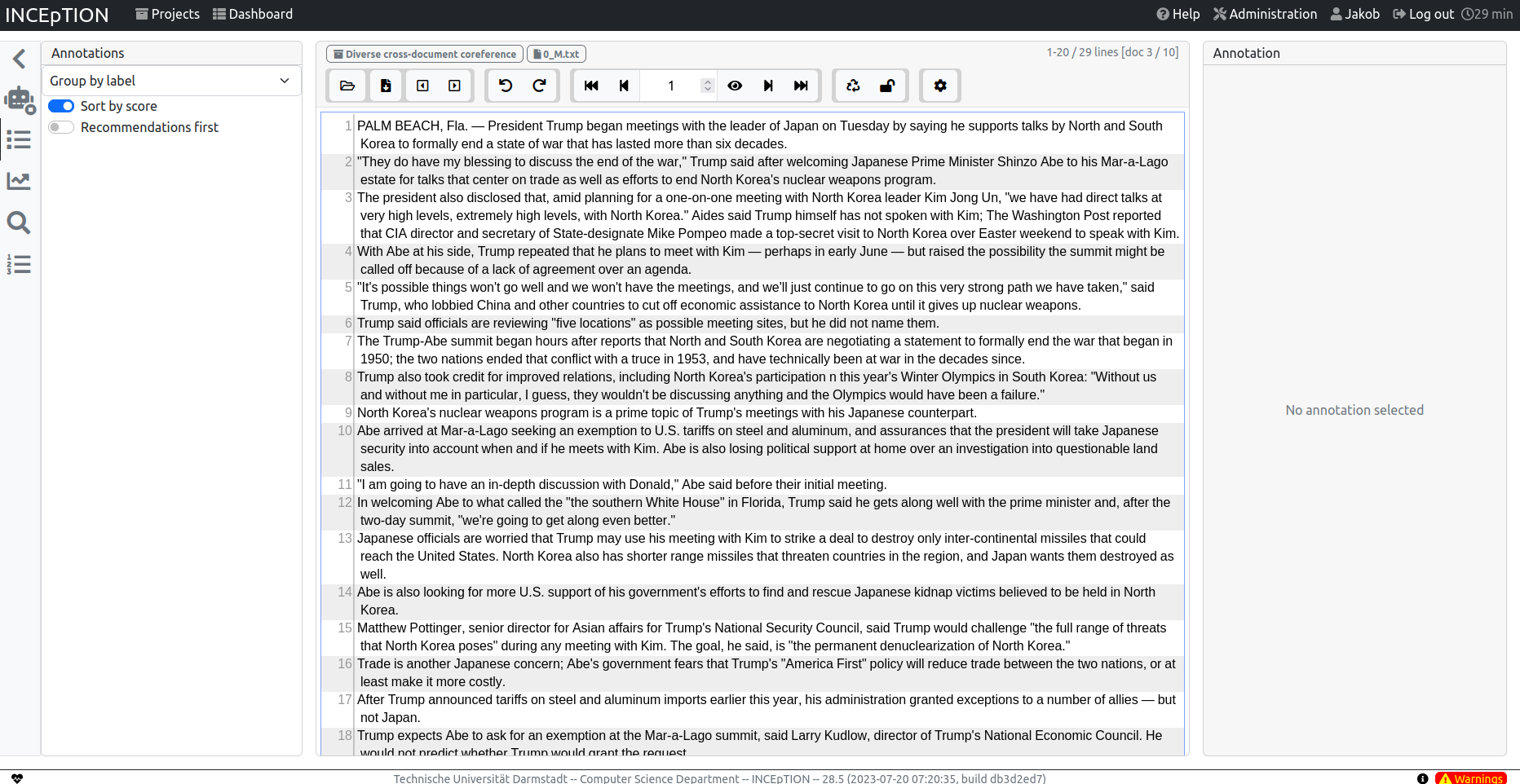}
\caption{Screenshot of Inception window showing a not yet annotated document loaded into the annotation GUI.}
\label{fig:inc5_emtpy_doc}
\end{center}
\end{figure}

\subsection{User manual}
Inception offers a variety of functionalities of which only those relevant for our project are described here. For a full explanation of how to use Inception, please check the official documentation which can be accessed \href{https://inception-project.github.io/releases/29.1/docs/user-guide.html}{online} or from within the Inception GUI by clicking on "Help" in the right upper corner.
Every annotator's instance of Inception contains two basic layers of annotation. The first layer, called \textbf{Entity layer}, is triggered when a mention is marked by highlighting text with a simple press-hold-drag mechanism. This opens the layer's side panel. Here, annotators can fill in the Entity layer's three parameters: 
\begin{itemize}
    \item \textbf{Entity-type}: a drop-down list to select a mention's entity-type by clicking on or typing the type's abbreviation.
    \item \textbf{Global entity-name}: a mixture of free text-field and drop-down list to assign a global entity's name to a mention. If the name has already been used before, it can be selected as item from the drop-down list by again clicking or typing. If not, it can be freely typed which adds it as a new tag to the list.
    \item \textbf{Wikidata}: a search field to type the name of an entity and find its respective Wikidata URI.
\end{itemize}
\begin{figure}[!ht]
    \begin{center}
    \includegraphics[width=8cm]{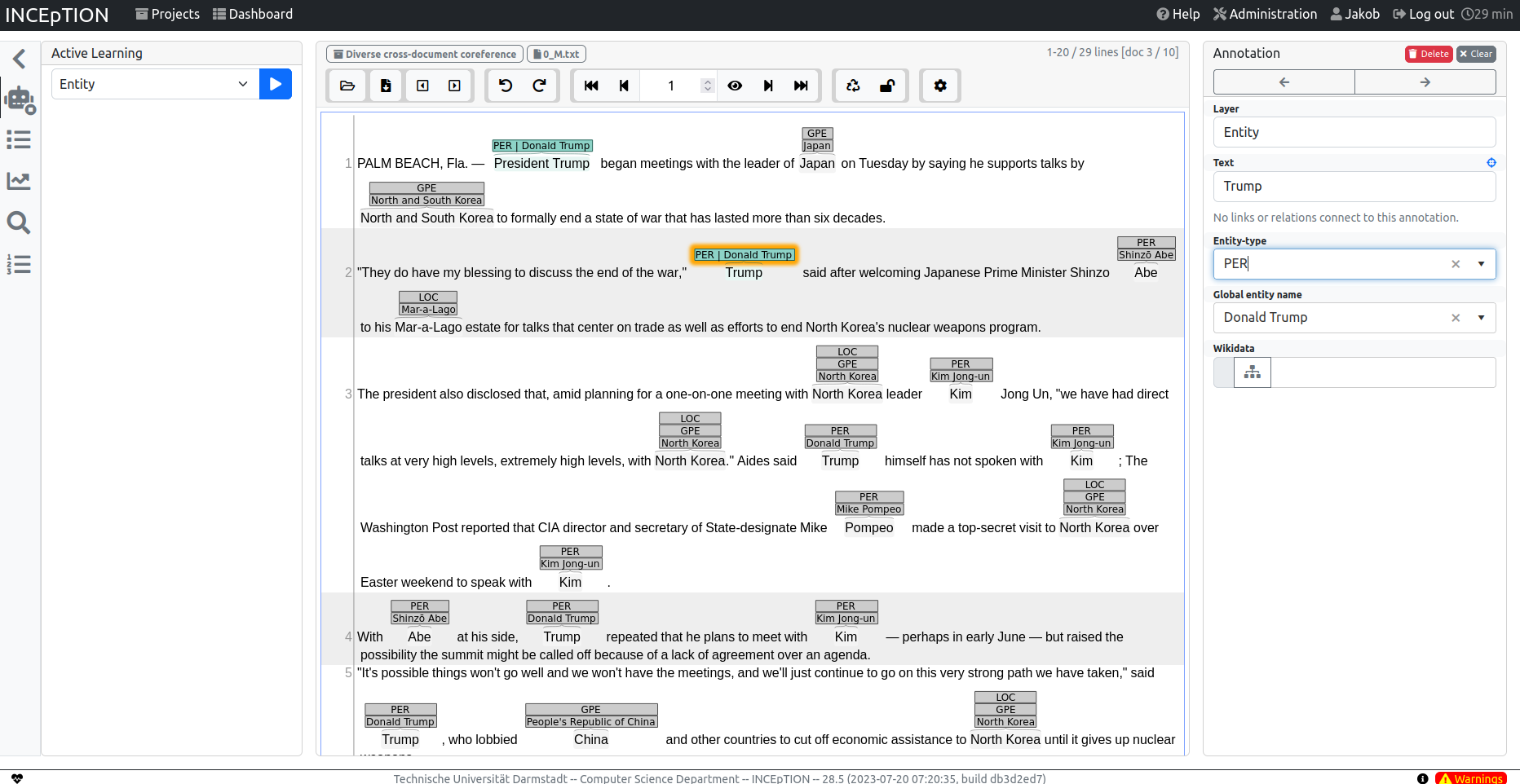}
    \caption{Annotating a mention of "Donald Trump": in the right panel, annotators can fill in values for the Entity layer's three parameters Entity-type, Global entity-name, and Wikidata. Automatically suggested annotations are displayed in gray boxes above the text rows.}
    \label{fig:inc7_annotation_mention}
    \end{center}
\end{figure}

The second layer, called \textbf{Relation}, is triggered when two already marked mentions are connected to each other, again simply by clicking and holding on one mention and dragging the mouse to the other mention. This layer only contains one parameter which is named \textbf{Label}. It is a drop-down list to select a relation-type for labelling the connection between both mentions. 

After the first annotations have been made, Inception starts to suggests spans and values for new annotations on the Entity layer. These suggestions are displayed in gray boxes. One click on a box accepts the suggestion and turns it into a proper annotation, a double-click denies the suggestion and makes the box disappear. 
\begin{figure}[!ht]
    \begin{center}
    \includegraphics[width=8cm]{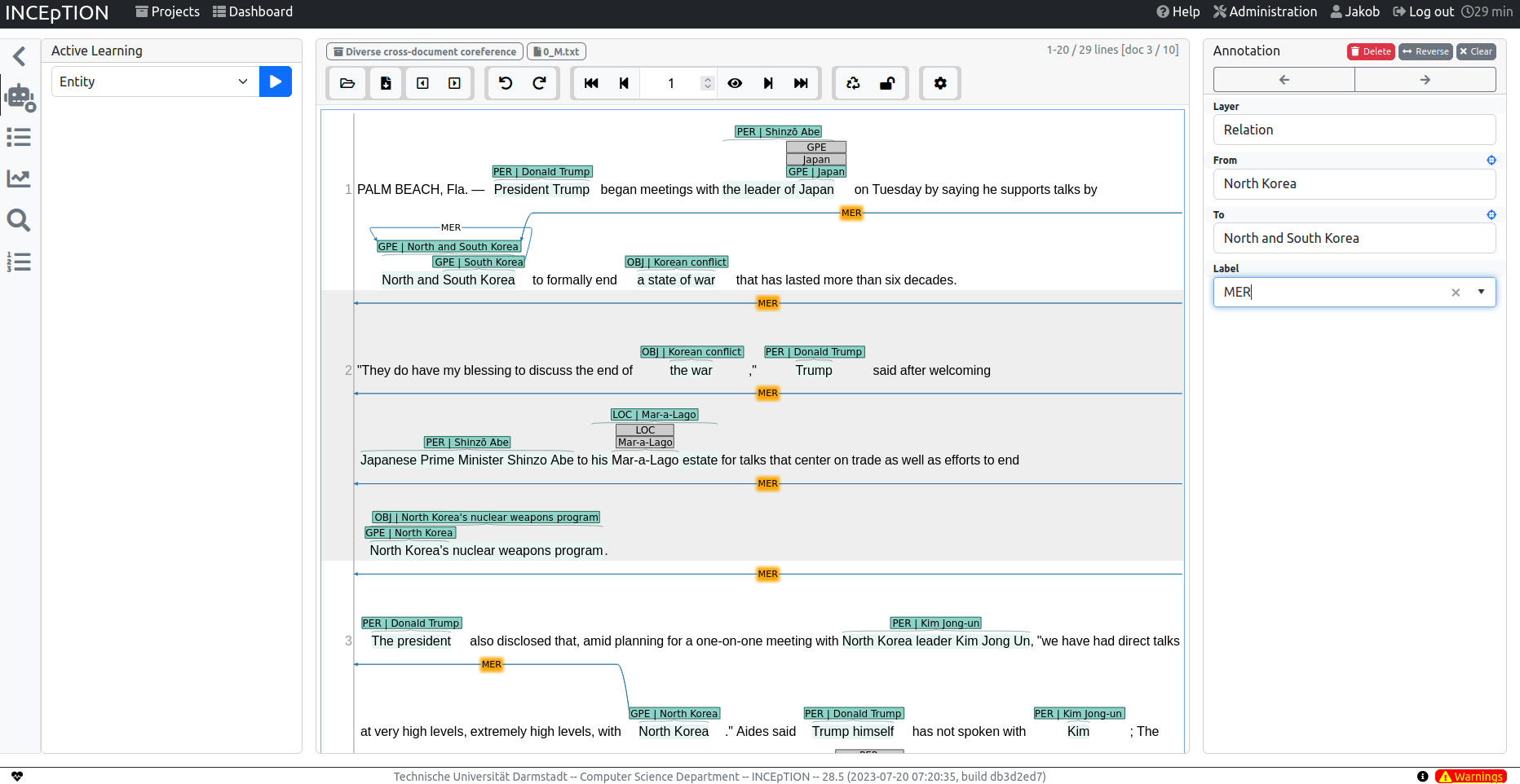}
    \caption{Annotating a relation between two mentions: the mention "North Korea" is connected to "North and South Korea" with a meronymy-relation (MER).}
    \label{fig:inc9_annotations_north_and_south_korea}
    \end{center}
\end{figure}

The GUI's upper panel is mostly for navigating through the document. However, it also contains a button for resetting the document by deleting all annotations made so far and a button in the shape of a padlock to mark the annotation process of the document as finished. This button should be pressed at the very end of the annotation, though it is advisable to first annotate each document before marking all of them together as officially finished. Clicking on the gear wheel opens up the GUI's style settings. Here, annotators have the option to adjust panels' margin sizes, the colouring of annotations, and how many text rows are to be displayed simultaneously. Annotations are saved automatically which is why there exists no saving button in the GUI.

\section{Annotation guidelines}
\label{sec:annotation}

Annotators will read each article three times and focus on a different annotation task in each pass: in the first pass, only read the text to get an overview of it. Do not make any annotations, yet. In the second pass, mark mentions with identity-relations, assign an entity to them and link them to Wikidata. In the third pass, annotate near-identity and bridging relations between mentions.

\subsection{First pass: get familiar with the text}

Read the entire text carefully. Try to already pay attention to what entities are mentioned, but do not annotate them, yet.

\subsection{Second pass: annotate mentions with identity-relations}

Read the text for a second time. Identify potential coreference candidates. Wherever a referent is referred to by at least two identical mentions, annotate these and all subsequent mentions respectively. Do this as follows:

\begin{itemize}
    \item First check if a candidate is markable: 
    \begin{itemize}
        \item In general, only \textbf{noun phrases (NPs)} are markable. This includes nominal phrases ("the president"), proper names ("Mr. Biden"), and quantifier phrases ("all member states").
        \item For reasons of efficiency, most pronominal NPs are excluded from annotation because they normally carry little variation with regards to how they are labelled \citep{zhukova2021towards}. However, certain types of pronouns can be included not as head, but as modifier for another NP, e.g. demonstrative pronouns ("this man") and reflexive pronouns ("the president himself").
        \item Numbers like currency expressions ("€2.3 billion") and percentages ("19\% of the votes") are included, but dates of any kind ("January 23", "1996", "this Sunday") are excluded for now.
        \item Given coreferential conjunctions that mention several entities at once and, syntactically, cannot be split ("North and South Korea"), first mark everything that could be extracted as single-entity mention separately (possible for "South Korea", but not for "North"), then mark the entire conjunction. Use a MER-relation to connect mentioned entities with the conjunction (see description of the MER-relation in subsection \ref{subsec:thirdpass}).
    \end{itemize}   
    
    \item Then check if the candidate you want to annotate is truly identical to other mentions of the same referent. To do so, compare it to the referent's most previous mention. In case no mention of the referent has been annotated so far, simply compare the two candidates triggering the annotation:
    \begin{itemize}
        \item Identity between two mentions means that both refer to the same entity in almost the same way. In comparison to the first mention, the second one may provide additional information about the referent or only highlight a subset of its attributes, but new and old attributes may not contradict each other \citep{recasens2010atypology}.
        \item When in doubt, ignore all modifiers and \textbf{focus on the heads} of both mentions to check if they are identical.
    \end{itemize}
    
    \item If the candidate is markable and identical to previous mentions, start your annotation. First, mark the mention:
    \begin{itemize}
        \item We annotate mentions with a \textbf{maximum span style}. This means that for each candidate, the NP's head and all of its pre- and post-modifiers are included in the annotation. More precisely, this includes articles ("a", the"), adjectives ("a worried president"), other NPs ("US president Joe Biden"), appositives  ("Joe Biden, president of the United States"), prepositional phrases ("demonstrators in front of the White House"), and relative clauses ("Biden, who was elected president in 2020") \citep{hirschman1998appendix}. Any punctuation or white space at the very beginning or end of the span are excluded.
        \item Additionally to maximum span style, we annotate with \textbf{nested style}, meaning a mention's span may overlap with or contain another mention. But remember not to mark any mention you discover, but only those who actually participate in coreference!
    \end{itemize}
        
    \item After selecting the correct span, assign an \textbf{entity-type} to a marked mention by choosing from the layer's respective drop-down list. We distinguish between the following entity-types: PER, ORG, GRP, GPE, LOC, OBJ.
    \begin{itemize}
        \item \textbf{Person} (PER): an individual actor.
        \item \textbf{Organization} (ORG): an official organization that is not government-related, e.g. "the WHO", "Fox News", "the opposition".
        \item \textbf{Group} (GRP): a group of individuals acting collectively or sharing the same properties, e.g. "demonstrators", "unemployed beneficiaries", "the two leaders".
        \item \textbf{Geo-political entity} (GPE): a state, country, province etc. that comprises a government, a population, a physical location, and a nation \citep{linguisticdataconsortium2008ace}. 
        This includes clusters of GPEs, e.g. "Eastern Europe" or "the Arab League". Governmental organizations or locations that represent an entire GPE are also marked as GPE, e.g. "the US government", "US officials", "the Biden administration", "Washington", "the White House". 
        \item \textbf{Location} (LOC): a physical location that is not a GPE, e.g. "Los Angeles". This includes mentions like "Germany" or "the White House" when referred to not in a political way, but with a focus on its geographic, cultural, architectural and other locality attributes. Be aware that two mentions with the same textual representation but different entity-types are not to be marked as identical! Instead, most of such cases would imply a MET-relation.
        \item \textbf{Object} (OBJ): an object or other concept that is mentioned, e.g. "Biden's hands", "a submarine", "the results". However, objects are static concepts. Do not confuse them with NPs that express events or other changes of state ("election", "negotiations", "Biden's statement") which we do not annotate!
    \end{itemize}
    \item Now it is time to assign the mention to an entity cluster. With this step, you create or extend a local coreference chain. At the same time, you link it with corresponding discourse entities across documents and globally with its actual referent.
    \begin{itemize}
        \item In case that, in the present document, you already have annotated previous mentions of the same entity, you will also already have created a local coreference cluster. The cluster will already be linked to a global discourse entity and to a referent. To assign the current mention to that cluster, select the global entity's name from the respective drop-down list. The Wikidata field can be left empty.\footnote{This is to save time. As the cluster will already be linked, assigning a Wikidata entry to every additional mention would be redundant work.}
        \item If, on the other hand, no previous mentions have been annotated, you are faced with two identical mentions you want to create a new local cluster of. To do this, first fill in the fields of the first mention. 
        \begin{itemize}
            \item Begin with the Wikidata field and type in the referent's name. Inception now looks for a suiting Wikidata entry and displays a drop-down list with the search results. Select the correct entry from that list. To enhance search results, try to look for the entity's most neutral name, ignoring articles. Sometimes it is easier to look for the entry on the Wikidata website itself and then copy its name into the field. If no Wikidata entry exists, leave the field empty.
            \item Assuming you have found a Wikidata entry, copy the text displayed in the Wikidata field into the Global entity-name field. By doing this, the name will automatically be added to the underlying tag set, meaning you will be able to select it from the drop-down list in subsequent annotations. However, if you have not found a Wikidata entry, copy the mention's text, again with maximum span style, into the Global entity-name field. Use this text as name for any following coreferential mentions. If the name has already been used for a semantically different entity in another document, add the document ID to the new name.\footnote{The following example illustrates this: let us assume you have annotated several mentions with the name "\textit{demonstrators}" in a previous document. Now, while annotating document "0\_L", you are faced with an entity that would also have to be given the Global entity-name "\textit{demonstrators}", although it refers to a semantically different group of people. In this case, do not change your annotations of the previous document, but do use the Global entity-name "\textit{demonstrators0\_L}" in the current document.}
        \end{itemize}
        \item Now turn to the second mention and annotate it based on the previous one. That is, assign the Global entity-name while leaving the Wikidata field empty.
    \end{itemize}

\end{itemize}

\subsection{Third pass: annotate mentions with different relations}
\label{subsec:thirdpass}
Read the text for a third time. Wherever you see two mentions connected through a near-identity relation, make a respective annotation:
\begin{itemize}
    \item For every new mention that has not been marked in the second pass already, check if it is markable and annotate it with its correct span and entity-type as described above. However, leave the Global entity-name and Wikidata field empty.
    \item When both mentions are marked with the correct span and entity-type, connect them with one of the following \textbf{near-identity relation-types}: MET, MER, CLS, STF, DEC, BRD \citep{recasens2010atypology, spala2019deft, clark2004changing, nedoluzhko2009coding}.
    \begin{itemize}
        \item \textbf{Metonymy} (MET): In a MET-relation, in comparison to its antecedent, an anaphor highlights different facets of an entity. This includes facets like:
        \begin{itemize}
            \item a certain role or function performed by an entity. Consider example (\ref{MET-1}).
            \begin{quote}
            \ex{MET-1}(\ref*{MET-1}) "Although \textit{Biden} is \textit{head of the Democrats}, he is also \textit{president of all Americans}."
            \end{quote}
            Assuming "\textit{Biden}" has already been annotated as part of a respective cluster in the second pass, "\textit{head of the Democrats}" and "\textit{president of all Americans}" would now be connected to "\textit{Biden}" with a MET-relation. However, in this example, it is the juxtaposition of both roles in particular that makes this a case of metonymy. In a more regular context, naming one of these roles alone could be annotated in the second pass as identical mention, instead.
            
            \item a location's name to refer to an associated entity, e.g. "\textit{Washington}" as metonym for "\textit{the US government}", "\textit{China}" for "\textit{the Chinese government}", "\textit{Silicon Valley}" for "\textit{the Tech industry}".
            
            \item an organization's name to refer to an associated place, e.g. a bank's name like "\textit{ECB}" to refer to the building that contains that bank's headquarters.
            
            \item different forms of realization of the same piece of information, like in example (\ref{MET-2}), where the same content is manifested once as audible speech and once as written text.
            \begin{quote}
            \ex{MET-2}(\ref*{MET-2}) "Though it is questionable whether he had actually written \textit{the piece} himself, Macron gave \textit{a truly brilliant speech} this afternoon."
            \end{quote}
            
            \item representation, where one mention is a picture or other representation of an entity, as already seen in example (\ref{near-ID-3}).
            \begin{quote}
            (\ref*{near-ID-3}) “The AfD is circulating \textit{a photo of Angela Merkel with a Hijab}, although \textit{Merkel} never wore Muslim clothes.”
            \end{quote}
            
            \item other facets, since this is no exhaustive list and metonymy is a dynamic phenomenon. 
            \item given two ID-clusters that are metonymous to each other (e.g. several mentions of "the US president" and several mentions of "the White House" which often participate in metonymy together), do not connect every single mention of the latter to a mention of the former, but only do this for the latter's first truly coreferential mention.
        \end{itemize}
        
        \item \textbf{Meronymy} (MER): A MER-relation between two mentions indicates that:
        \begin{itemize}
            \item one mention is a constituent part of the other in whatever direction, as in example (\ref{MER-1}).
            
            \begin{quote}
            \ex{MER-1}(\ref*{MER-1}) "\textit{President Biden} expressed his concern about the ongoing ... '\textit{The US government} will not ...', he stated."
            \end{quote}
            
            \item one mention refers to an object which is made of the stuff which the other mention refers to.
            \begin{quote}
            \ex{MER-2}(\ref*{MER-2}) "The duty on \textit{tobacco} has risen once again, making \textit{cigarettes} as expensive as never before."
            \end{quote}
            \item both mentions refer to overlapping sets.
             \begin{quote}
            \ex{MER-3}(\ref*{MER-3}) "\textit{AfD supporters} demonstrated in front of the Reichstag this morning. Among \textit{the crowd} was ..."
            \end{quote}
            
            \item finally, a MER-relation can be used to specify entities mentioned in syntactically non-dividable conjunctions. Given such a conjunction, as "North and South Korea" in example (\ref{CONJ-1}), mark "South Korea" separately as it can be treated as independent noun phrase. The adjective phrase "North", however, cannot be marked. Instead, mark the entire conjunction and connect "South Korea" to it with a MER-relation (illustrated by the dotted underlining). Do the same for the first full mention of "North Korea" that follows in the text. If none follows, use a previous mention or, if there is none, ignore the "North"-mention.
            
            \begin{quote}
            \ex{CONJ-1}(\ref*{CONJ-1}) "\textit{North and \dotuline{South Korea}} have resumed negotiations ... \textit{North Korea} seems ..."
            \end{quote}
            
        \end{itemize}
        \item \textbf{Class} (CLS): a CLS-relation indicates an 'is-a' connection between two mentions. One mention thus belongs to a sub- or superclass of another. 
        \begin{quote}
        \ex{CLS-1}(\ref*{CLS-1}) "In way, \textit{Trump} only seized the opportunity. This is what \textit{skilled politicians} do."
        \end{quote}
                
        \item \textbf{Spatio-temporal function} (STF): a mention refers to an entity that deviates in place, time (\ref{near-ID-2}), number, or person (\ref{STF-1}).
        \begin{quote}
        (\ref*{near-ID-2}) "Even if \textit{the young Erdogan} used to be pro-Western, \textit{Turkey's president} nowadays often acts against Western interests."
        
        \ex{STF-1}(\ref*{STF-1}) "A historic meeting: \textit{a pope} and \textit{a pope} shaking hands."
        \end{quote}
        
        \item \textbf{Declarative} (DEC): where two mentions X and Y are connected through verbal phrases like "X seems like Y", "stated that X was Y", "declared X Y", or other declarations as in (\ref{DEC-1}), they can be connected with a DEC-relation. 
        \begin{quote}
        
        \ex{DEC-1}(\ref*{DEC-1}) "In his speech, he also spoke about \textit{North Korea} and called it \textit{a fundamentally barbaric nation}."
        \end{quote}
        
        The DEC-relation thus includes definitions and descriptions of entities. This is especially the case when declarative clauses are used within quotes. However, when value-free declarative clauses like "X is Y" are used as quasi objective specifications of an entity, they might indicate an identity relation, instead. The same structure might be used to assign a super-class to the entity, making it a CLS-relation.

    \item \textbf{Bridging} (BRD): for reasons of simplicity, we have included BRD in our subsumption of different relation-types under the term of near-identy. Despite of that, BRD is actually a separate phenomenon from both identity and near-identity. BRD connects two entities that are mostly independent of each other while nonetheless, the existence of one can be inferred by the existence of the other \citep{clark1977comprehension}. Technically, the BRD-relation could be used to mark all sorts of ontological connections between entities. This is not the purpose of this annotation scheme, though. Instead, we use BRD only where the mention of one entity influences the depiction of an associated entity or where one entity is modified by a possessive pronoun that refers to another entity. Example (\ref{BRD-1}) illustrates both use cases:
    \begin{quote}
    \ex{BRD-1}(\ref*{BRD-1}) "Unlike \textit{Queen Elizabeth}, \textit{Charles} has not been shy about promoting \textit{his political views}."
    \end{quote}
    Here, the NP "\textit{his political views}" contains a modifying possessive pronoun, which is why it is to be annotated as bridging to "\textit{Charles}". Additionally, the mention "\textit{Charles}" can only be interpreted correctly as referring to Charles III (and not any other Charles) by its juxtaposition with the NP "\textit{Queen Elizabeth}". Hence "\textit{Charles}" is to be annotated as bridging to "\textit{Queen Elizabeth}".
    \end{itemize}
    \item Deciding on what relation-type to choose can be difficult. When in doubt, follow these general guidelines:
    \begin{itemize}
        \item use an identity relation rather than a near-identity relation (especially DEC).
        \item when having to choose between near-identity relations, use MET rather than MER.
        \item use MER rather than CLS.
        \item use CLS rather than DEC.
        \item use any near-identity relation that is not BRD rather than BRD.
    \end{itemize}
    \item When annotating near-identity and bridging, always connect an anaphoric mention to the nearest possible antecedent. But remember that antecedents normally appear before an anaphor. Only if necessary you may connect a mention to a subsequent expression (making their relation cataphoric). 

\end{itemize}

\section{Conclusion and future work}
Our proposed annotation scheme covers a multitude of coreferential relations. It gives a detailed explanation of how to mark coreferential mentions across documents, assign entity-types and names to them, connect them with each other, and link them to the Wikidata knowledge graph. The scheme thus represents a significant step toward more accurately capturing the complexities of coreference use. It furthermore provides a valuable resource for researchers both in the field of coreference resolution and media bias by word-choice and labelling. Having said that, our scheme leaves room for possible extensions to further advance research in those domains. First, the annotation of events could be included in our scheme. An interesting question that arises is whether the relation-types as outlined here could be applied not only to entities, but to events all the same. A second possible extension would be to include a layer of media bias annotation to the scheme, enabling a direct comparison of diverse coreference usage and media bias by word-choice and labelling. Both proposed extensions could be easily added on top of our scheme.
Having said that, the present form of our scheme already addresses many of the complexities of diverse cross-document coreference and offers a roadmap for capturing nuanced linguistic relationships, ultimately advancing our understanding of language and discourse in digital print media.

\section{Acknowledgements}

Many thanks to my project supervisor Anastasia Zhukova who never became tired of my many questions and always knew how to help me out with good advise whenever I felt stuck.

\section{Bibliographical References}\label{reference}
\bibliographystyle{lrec-coling2024-natbib}
\bibliography{ms}

\end{document}